\documentclass[conference]{IEEEtran}
\IEEEoverridecommandlockouts

\usepackage{cite}
\usepackage{amsmath,amssymb,amsfonts}
\usepackage{algorithmic}
\usepackage{graphicx}
\usepackage{textcomp}
\usepackage{xcolor}
\usepackage{hyperref}
\hypersetup{hidelinks}

\usepackage{subfigure}
\usepackage{lipsum}  
\def\BibTeX{{\rm B\kern-.05em{\sc i\kern-.025em b}\kern-.08em
    T\kern-.1667em\lower.7ex\hbox{E}\kern-.125emX}}
\begin{document}

\title{COMB: Common Open Modular robotic platform for Bees\\
\thanks{\textsuperscript{1}Department of Computer and Information Science, University of Konstanz, Germany \par  
\textsuperscript{2}Centre for the Advanced Study of Collective Behaviour, University of Konstanz, Konstanz, Germany\\
{\tt{ [pranav.kedia]@uni-konstanz.de  }}\par
\textsuperscript{3}Dahlem Center for Machine Learning and Robotics, Freie Universität Berlin, Berlin, Germany\\
{\tt{ [marie.messerich, tim.landgraf]@fu-berlin.de  }}\\
Corresponding author: Pranav Kedia\\
}
}

\author{Pranav Kedia\textsuperscript{1,2,3},  Marie Messerich\textsuperscript{3} and Tim Landgraf\textsuperscript{3}%
      
}%

\maketitle

\begin{abstract}
Experimental access to real honeybee colonies requires robotic systems capable of operating within limited spatial constraints, tolerating hive-specific fouling and environmental conditions, and supporting both sensing and localized actuation without frequent hardware redesign. This paper introduces \textit{COMB}, a compact, open-source, modular mechatronic platform designed for in-hive experiments within standard observation-hive frames. The platform integrates a XY positioning stage, a Movable Access Window (MAW) for sealed tool access through the hive boundary, interchangeable payload modules, and an embedded control architecture that enables repeatable trajectory execution and signal generation. The platform's capabilities are demonstrated through three representative modules: a biomimetic dance-and-signaling payload, a close-range comb scanner, and an electromagnetic wing actuator for localized oscillatory stimulation.\\
This paper details the hardware and software design of COMB, outlines its operational capabilities, and describes the supporting infrastructure for conducting real-world in-hive experiments. The platform is characterized in engineering terms through tracking waggle-trajectory executions, performing multi-image stitching for repeated comb mosaics, and conducting video-based spectral analysis of the wing actuator. These results position COMB as a reusable experimental robotics platform for controlled in-hive sensing and actuation, and as a compact, generalized successor to earlier task-specific honeybee robotic systems.
\end{abstract}


\section{Introduction}
Honeybees (\textit{Apis mellifera}) are a cornerstone model for studying social communication, learning and memory, navigation, and collective behavior. While colonies are typically housed in full-size hives, laboratory investigations often use observation hives, most commonly two comb frames stacked with glass windows, to enable direct, long-duration viewing of natural behavior. Some experiments further require unobstructed access to the comb surface for manipulation or high-fidelity sensing (e.g., acoustic or vibrational recordings); in these cases, groups sometimes heat the room and work with open observation hives to maintain colony homeostasis during interventions.\par

Within this experimental landscape, the waggle dance remains a central communication behavior and a touchstone for mechanistic inquiry. Since von Frisch’s seminal work \cite{vonFrisch1965tanzsprache}, the dance has inspired decades of research into how foragers encode and nestmates decode directional and distance information under the low-light, crowded conditions of the hive. The dance comprises a central waggle run and return loops on the comb; its duration and orientation reflect the resource direction, distance, and quality \cite{seeley_dancing_2000}.  Multiple sensing modalities contribute to how followers perceive the dance when vision is limited \cite{landgraf_analysis_2011}. Early biomimetic experiments with mechanical “dummy” bees demonstrated uncoupling cues that in live animals are hard to control \cite{michelsen_how_1992}, yet important nuances of the emitted signals and the decoding strategies of followers remain unresolved \cite{d_bierbach_dancing_2018}.\par

Animal–robot interaction offers a principled path to address these questions with precision while retaining ecological validity. In so-called animal-robot interaction systems (ARIS), researchers embed engineered agents into animal groups to inject, modulate, or replace specific cues and measure closed-loop responses in real-time \cite{landgraf2021animal,romano_review_2019,romano2024animal,schmickl2024robots}. Across taxa, including fish, cockroaches, and even plants, well-designed robotic conspecifics have shaped individual and group dynamics \cite{g_sempo_social_2007,cazenille_how_2018,romano2024robot,barmak_robotic_2023}. In honeybees, a robotic waggle dancer has shown that followers can attend to artificial dances and subsequently depart in the advertised direction \cite{landgraf_biomimetic_2010,d_bierbach_dancing_2018}, albeit with inconsistent elicited dance-following behavior that highlights remaining knowledge gaps. \par
Beyond waggle dance robots, other state-of-the-art research explored various ways of bee-robot interaction. One way is to put electronic and robotic capabilities inside the hive environment to modulate various conditions and stimuli that produce a bee response. For instance, recent work showed a “robotic honeycomb” \cite{barmak_robotic_2023} equipped with thermal sensors and heat-producing elements, to observe the colony and provide real-time feedback. In a hive of thousands of bees, this device monitored and observed thermal patterns of bees and influenced the colony’s clustering behavior by delivering calibrated heat signals. Another study investigated the honeybee shaking signal \cite{koenig_artificial_2020}, a communication behavior wherein a worker bee vibrates another individual to increase its activity. By designing a motorized shaker that mimicked the vibrational frequency and duration of natural signals, the researchers demonstrated that bees(both workers and drones) increased movement in response and engaged in social behaviors such as grooming and feeding. Lastly, the observation robot \cite{ulrich2024autonomous} with a moving camera was used to observe and track bees for an extended period of time. The system is crucial to detect behavioral characteristics of the queen, workers, and brood in the comb. This was a follow-up study of the scanning ability of COMB.   Together, these advances represent the state of the art in animal-robot interaction with bees. They not only enable researchers to study bees under controlled yet naturalistic conditions, but also point to practical applications like improving pollination or safeguarding colonies through technology.\par

Despite these achievements, most bee-robot systems are custom-built for specific laboratories, hive configurations, and research questions. The hive itself presents additional challenges, such as thermal regulation, airflow, propolis buildup, biofouling, and mechanical durability during overcrowding. These factors complicate the reuse of these systems. What is currently missing is a compact and reusable in-hive platform that provides standardized mechanical access to the comb while supporting interchangeable sensing and actuation modules. Such a platform should fit common observation-hive formats, maintain a barrier between the hive interior and external mechanics, provide repeatable positioning over the comb surface, and expose a shared hardware and software interface for diverse payloads. Addressing this gap would reduce duplication of effort across laboratories and make robotic honeybee experiments easier to reproduce and extend.\par

\begin{figure}[htbp]
\centering
\subfigure{\includegraphics[width=0.9\columnwidth]{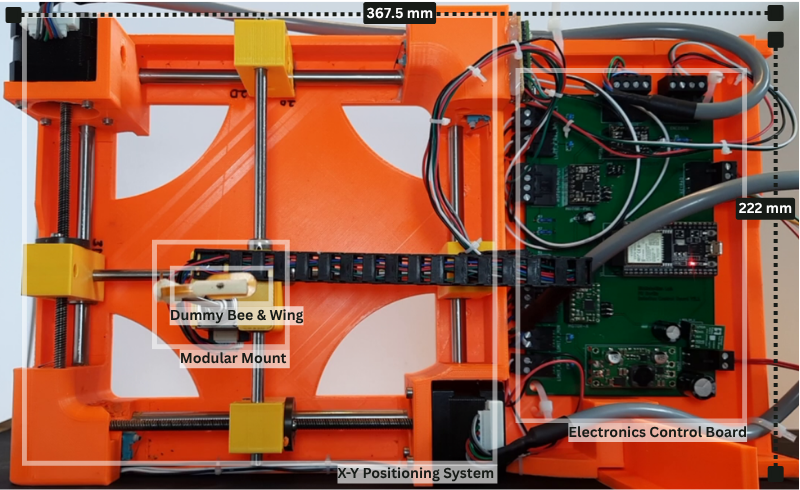}} 
\caption{COMB in dance-signaling configuration}
\label{fig:figdance}
\end{figure}

\begin{figure}[htbp]
\centering
\subfigure{\includegraphics[width=0.8\columnwidth]{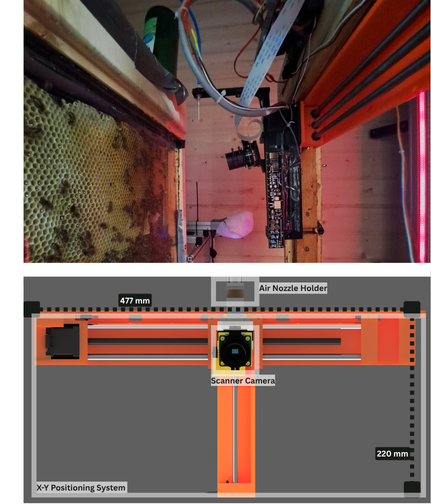}}
\caption{COMB configured in comb-scanning mode. Top: in-hive deployment in front of a two-frame observation hive. Bottom: front-view CAD render.}
\label{fig:figscan}
\end{figure}

Against this backdrop, we introduce COMB (Figs.~\ref{fig:figdance} and~\ref{fig:figscan}), a compact, open-source, modular mechatronic platform for experimental access to honeybee colonies. The system is sized for standard Deutsch-Normalmaß (DNM) frames and combines a two-dimensional positioning stage, a sealed Movable Access Window (MAW), embedded control electronics, and a payload interface for interchangeable modules. In contrast to earlier single-purpose systems, COMB is intended as a general in-hive research platform that supports both observation and interaction tasks within a common mechanical and control architecture. In this sense, COMB can be viewed as a smaller and more general successor to earlier task-specific honeybee robotic systems, including previous waggle-dance platforms developed for controlled signaling experiments \cite{landgraf_biomimetic_2010}. We shift the emphasis from a single dedicated robotic behavior toward a reusable mechatronic base that can host multiple sensing and actuation modules in the same hive-compatible form factor.\par

\section{Contributions}
In summary, this paper makes four main contributions. First, it presents a compact modular mechatronic platform for in-hive experiments in standard DNM-sized observation hives, combining a shared motion stage, embedded control architecture, and payload interface. Second, it introduces a Movable Access Window (MAW) that enables repeated tool access through the hive boundary while preserving visibility and reducing direct interference with the colony interior. Third, it demonstrates the platform through three representative modules: a comb scanner, a biomimetic dance-and-signaling payload, and an electromagnetic wing actuator for localized oscillatory stimulation. Fourth, it characterizes the platform in engineering terms through trajectory tracking, image mosaicing, spectral analysis of the flapping actuator, and seasonal in-hive deployment observations. The open-source release of the hardware and software stack is intended to support reuse and reproducibility.

\section{Setup and Hardware System Design}

\subsection{Design Requirements and System Overview}
The platform was designed to satisfy four practical requirements for in-hive robotics. First, it had to fit within the envelope of a standard-sized observation-hive frame without obstructing colony visibility. Second, it had to support repeatable positioning of tools over the comb surface for both imaging and localized actuation. Third, it had to preserve a barrier between the hive interior and external mechanics in order to limit airflow disruption, fouling, and accidental bee access to the drive system. Fourth, it had to support interchangeable payloads without redesigning the full mechanical or electronic stack. These requirements motivated a common architecture consisting of a planar positioning unit, a sealed access interface, a shared control board, and hot-swappable center modules.\par

\subsection{Mechanical Setup}
The mechanical core of COMB is a compact XY positioning stage sized to the DNM frame geometry used in our observation hives. DNM frames are commonly used by German beekeepers, thereby enabling COMB to integrate seamlessly with traditional beekeeping equipment. The stage carries the active payload on a front-facing carriage aligned with the comb surface, allowing the same base mechanism to be used for dance-signaling (Fig.~\ref{fig:figdance}), comb-scanning (Fig.~\ref{fig:figscan}),  or other manipulator modules like the flapper (Fig.~\ref{fig:figprobe}). The layout keeps most motors and electronic components outside the immediate hive-facing workspace, while the moving carriage establishes a consistent interface plane for all end effectors. This arrangement minimizes repeated redesign of the mechanical frame when changing experimental modules.\par

COMB is designed around the \emph{DNM} frame dimensions,  Moreover, adapting the core assembly to other standard hive sizes, such as the \emph{Zander} format (frequently used in Austria) scanning robot (Fig.~\ref{fig:figscan}), requires minimal modification and demonstrates the platform’s flexibility. The mechanical design can be viewed as a more compact and generalized evolution of earlier task-specific honeybee robotic systems, retaining comb-surface access while broadening the range of supported payloads.\par

\subsection{Movable Access Window}
\begin{figure}[htbp]
\centering
{\includegraphics[width=0.8\columnwidth]{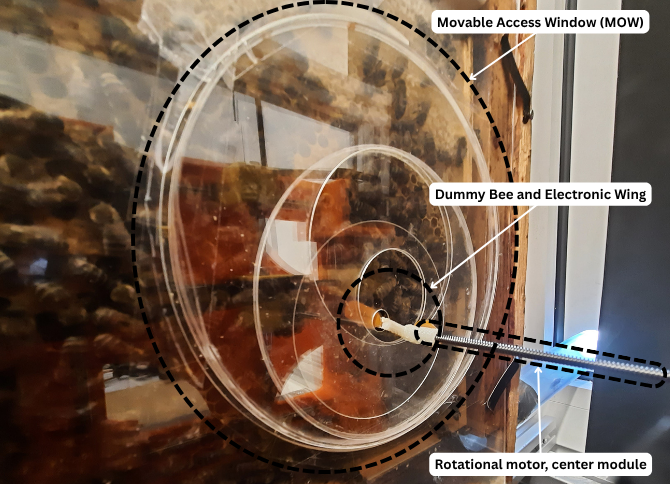}}
\caption{Movable Access Window (MAW) inserted with dance-signal bee module}
\label{fig:figmow}
\end{figure}

Observation–hive experiments require a barrier that protects both colony homeostasis and research hardware while still allowing precise, repeatable access to the comb surface. Bees readily deposit propolis on moving parts, react to drafts, and are sensitive to temperature gradients; at the same time, end-effectors (e.g., a dummy bee dancer, or flapper probe) must pass into the hive without inviting excessive airflow or outward bee traffic. To meet these constraints, we designed a \textit{Movable Access Window (MAW)}—a transparent, rotary insert laminated to the thickness of the hive glazing—that preserves thermal and olfactory isolation while providing a clear, sealed feed-through for tools (Fig.~\ref{fig:figmow}).\par

\paragraph*{Form factor and materials}
The MAW is integrated into a 3\,mm transparent polycarbonate panel cut to the hive's inner frame dimensions \(L\times B\). A centered circular aperture of radius
\[
R = \frac{B}{2} - c
\]
is cut into the panel, where \(c \ge 10\,\mathrm{mm}\) provides a margin against cracking and mounting loads. Polycarbonate was chosen for its higher impact resistance and durability compared with acrylic.\par

\paragraph*{Rotary insert and seal geometry}
A three-layer laminate (each \mbox{1\,mm} thick) forms a \mbox{3\,mm} moving insert that drops into the aperture: top and bottom \emph{discs}, each of outer radius \(R-\varepsilon\) with an inner circular cutout of radius $r$, and a middle \emph{ring} of thickness \(s\) that creates a spacer at the edge. The outer radius of the insert is \(R-\varepsilon\), where \(\varepsilon\) (\(\sim\!0.2{-}0.4\,\mathrm{mm}\)) is the running clearance for smooth insertion. The relationship
\[
r \;=\; R - s - \varepsilon
\]
ensures that the overlap \(s\) sets the seal path length while the clearance \(\varepsilon\) limits leakage and rattle. Lastly, a \emph{smaller disc} of radius $(2*r )$ with a central cutout of radius $5mm$ which is the shaft radius of the bee dummy. The additional r measure of radial overlap ensures a snug fit against the inner edges of the two larger layers.\par

\paragraph*{Assembly and performance}
Once bonded, the laminate forms a transparent rotary insert that fits into the main panel aperture and acts as a sealed interface for the robot's active components. In long-term use, the MAW minimized drafts and heat loss, tolerated propolis buildup, and could be restored with periodic isopropyl cleaning. This reduced direct contact between bees and moving parts, lowering contamination and improving mechanism longevity.

\subsection{Electronics and Embedded Control}
The electronics of \textit{COMB} were designed as a shared embedded control layer for the motion stage and interchangeable payload modules. At the core of the system is an ESP32-based controller running the platform logic for motion execution, peripheral coordination, and local user interaction. This architecture allows the same electronics stack to support multiple operating modes, including programmed waggle-trajectory playback, close-range comb scanning, and localized oscillatory stimulation, without redesigning the control hardware for each experiment.\par

The XY positioning stage is actuated through dedicated stepper motor drivers, with one driver assigned to each translational axis. When required by the installed module, an additional actuator channel is used for payload-specific motion, such as dummy orientation or wing-related excitation. End-stop switches define the physical workspace and provide a basic hardware safety layer during manual positioning and automated trajectory execution. In this way, the electronics support both repeatable scripted motion and operator-guided adjustment during calibration, deployment, and in-hive maintenance.\par

To simplify operation in laboratory and hive environments, the platform includes a compact keypad-based local interface. This allows the operator to jog the stage, enable or disable motion, switch between operating modes, and trigger predefined routines without relying on a continuously attached host computer. At the same time, the embedded controller synchronizes stage motion with payload-level actuation, enabling coordinated behaviors such as trajectory-following with simultaneous signal generation or scanning. This combination of embedded control, local usability, and synchronized module operation is central to the platform’s modularity, since it allows different sensing and actuation payloads to be treated as extensions of the same common robotic base rather than as separate one-off systems.\par

\subsection{Representative Payload Modules}
To demonstrate the flexibility of the common platform, we implemented three representative payload modules that span the two main roles targeted by \textit{COMB}: observation of the colony and controlled physical interaction with it. These are interchangeable end effectors mounted on the same hive-compatible robotic base, showing how one shared platform can support multiple experimental tasks.\par

\begin{figure}[htbp]
\centering
{\includegraphics[width=0.8\columnwidth]{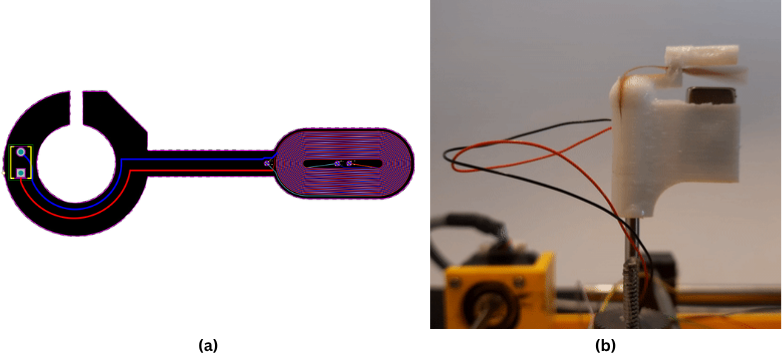}}
\caption{(a) Electronic Wing PCB, (b) The bee-dummy with the flapping wing}
\label{fig:figflap}
\end{figure}

The first module is a biomimetic dance-signal payload designed to reproduce programmed motion patterns on the comb surface. In this configuration, the carriage carries a dummy bee end effector (Fig.~\ref{fig:figflap}(b)) whose motion is driven by the underlying XY stage. The module is primarily intended for executing predefined waggle-like trajectories and other localized movement patterns relevant to honeybee interaction experiments. \par

The second module is a close-range comb scanner intended for tiled imaging of the hive surface. Mounted on the XY carriage, the scanner follows programmed raster-like trajectories and acquires overlapping images of local comb regions, which can then be stitched into larger mosaics. By repeating the same scan path across time, the module also supports timelapse-style observation of brood development, food storage, or local colony activity.\par

\begin{figure}[htbp]
\centering
{\includegraphics[width=0.9\columnwidth]{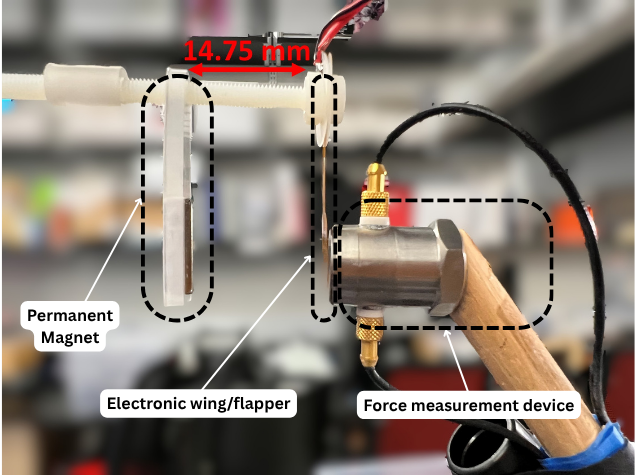}}
\caption{Electronic Wing Actuator separated from the permanent magnet, being used to provide shaking signals at the Nieh lab at UC San Diego}
\label{fig:figprobe}
\end{figure}

The third module is an electromagnetic wing/flapper actuator (Fig.~\ref{fig:figprobe}) used for localized oscillatory stimulation. This payload employs a flexible PCB-based coil (Fig.~\ref{fig:figflap}(a)) and permanent magnet arrangement to generate repeatable wing-like motion under electrical excitation. Compared with larger external mechanisms, it offers a compact and compliant way to deliver mechanical signals directly to bees. \par

Together, these modules demonstrate the intended scope of \textit{COMB}, a single mechatronic platform capable of supporting observation, biomimetic motion, and localized actuation within live honeybee colonies.\par

\section{Experiments and Results}
The \textit{COMB} system was evaluated as a mechatronic research platform. Accordingly, the experimental investigation was structured around four principal engineering questions: (1) the extent to which the platform is capable of reproducing programmed waggle-like trajectories with repeatable accuracy; (2) whether the scanner module facilitates tiled, high-resolution close-range comb imaging and subsequent reconstruction; (3) the ability of the oscillatory payload to generate controlled and quantifiable stimulation frequencies; and (4) the operability of the hive interface during extended seasonal deployment, particularly under conditions of propolis exposure.

\begin{figure}[htbp]
\centering
{\includegraphics[angle=180,width=0.6\columnwidth]{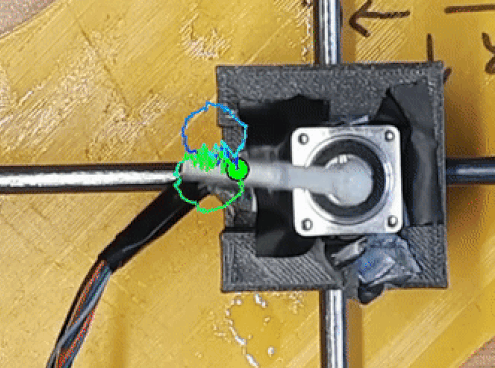}}
\caption{Tracked bee-dummy trajectory during waggle execution}
\label{fig:figtraj}
\end{figure}

\begin{figure}[htbp]
\centering
\subfigure[ \label{fig:XY_traj}]{
\includegraphics[width=\columnwidth]{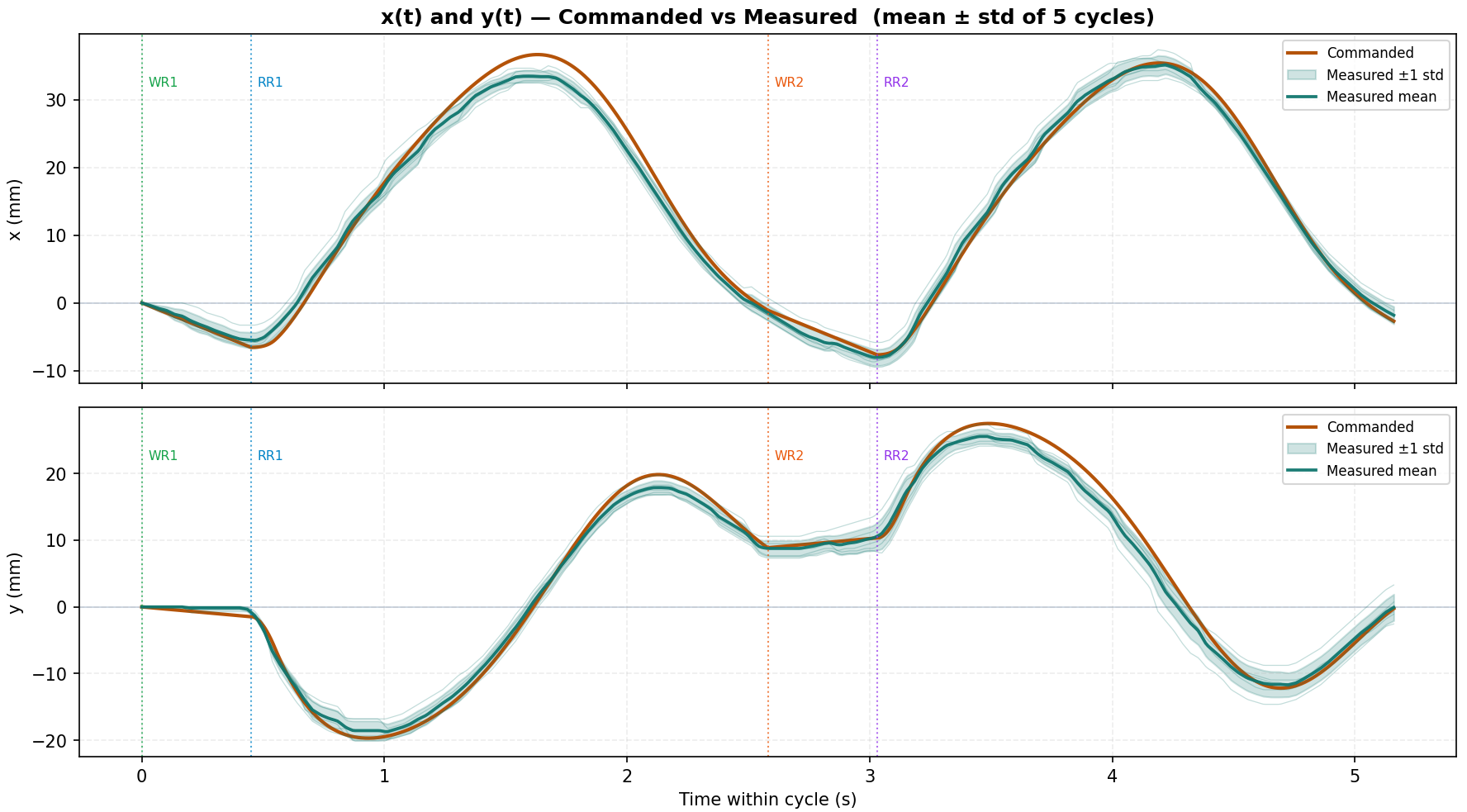}
}
\subfigure[ \label{fig:CTE_error} ]{\includegraphics[width=\columnwidth]{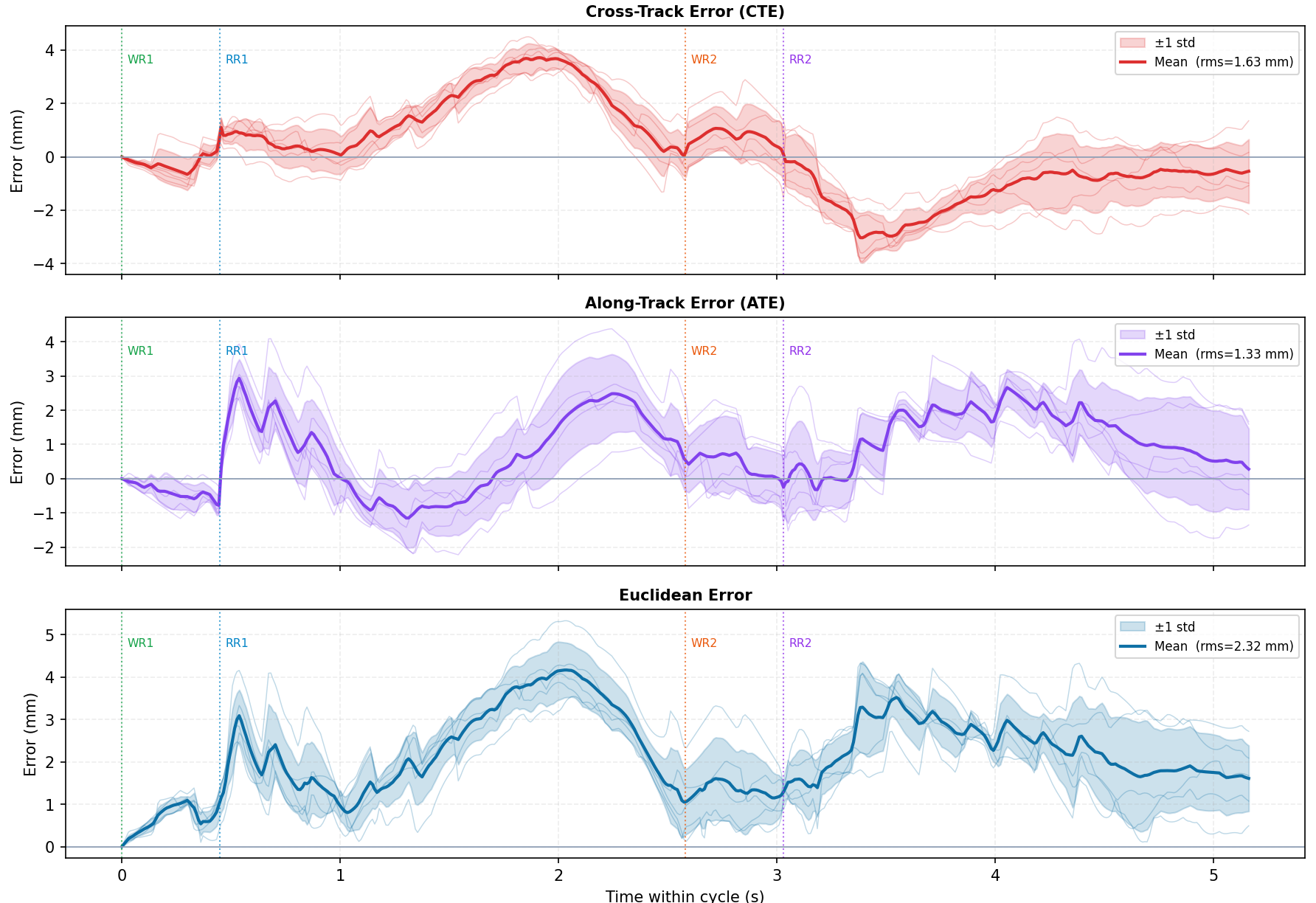}}
\caption{Trajectory tracking over five runs: (a) mean executed path in~x and~y, (b) tracking error decomposition.}
\end{figure}

\paragraph*{Waggle-Trajectory Tracking Performance} To characterize the motion accuracy of the system in dance mode, we analyzed five representative executions of the same programmed waggle trajectory using overhead video and an OpenCV-based tracking pipeline. A typical tracked dummy trajectory is shown in Fig.~\ref{fig:figtraj}. For each run, the bee dummy was tracked in image space, converted to physical units using a calibration factor of 5.48~px/mm, and temporally normalized to one cycle for comparison against the commanded path. The measured trajectories were then averaged across runs to obtain a mean executed path and its variability over time. The resulting average path and the associated error metrics are shown in Fig.~\ref{fig:XY_traj} and Fig.~\ref{fig:CTE_error}.\par
The measured and commanded trajectories exhibited close correspondence along both spatial axes. Across the five runs, the root-mean-square cross-track error (CTE) was 1.63~mm, with a maximum cross-track deviation of 3.71~mm. The corresponding along-track error (ATE) had an RMS value of 1.33~mm, while the overall Euclidean trajectory error was 2.32~mm RMS. Notably, errors were not uniformly distributed throughout the cycle; the largest deviations were observed near high-curvature transitions and at peak lateral excursions, whereas the central waggle segment was tracked with greater fidelity. These findings indicate that the primary limitation of the platform lies in curvature-dependent path following, rather than cumulative drift or synchronization loss. Collectively, these results demonstrate that \textit{COMB} is capable of executing programmed waggle-like motion with millimeter-scale repeatability across repeated trials.

\paragraph*{Comb Scanning and Mosaic Reconstruction} To validate the observation capability of the platform, we configured \textit{COMB} in scanner mode using a close-range imaging payload mounted on the same XY base. In this mode, the carriage follows programmed raster-like paths and acquires overlapping local images of the comb surface, which are later stitched into larger mosaics. The scanner is intended to support non-destructive imaging of brood, stores, and comb occupancy without changing the underlying robot body or hive interface.\par
\begin{figure}[htbp]
\centering
{\includegraphics[width=0.8\columnwidth]{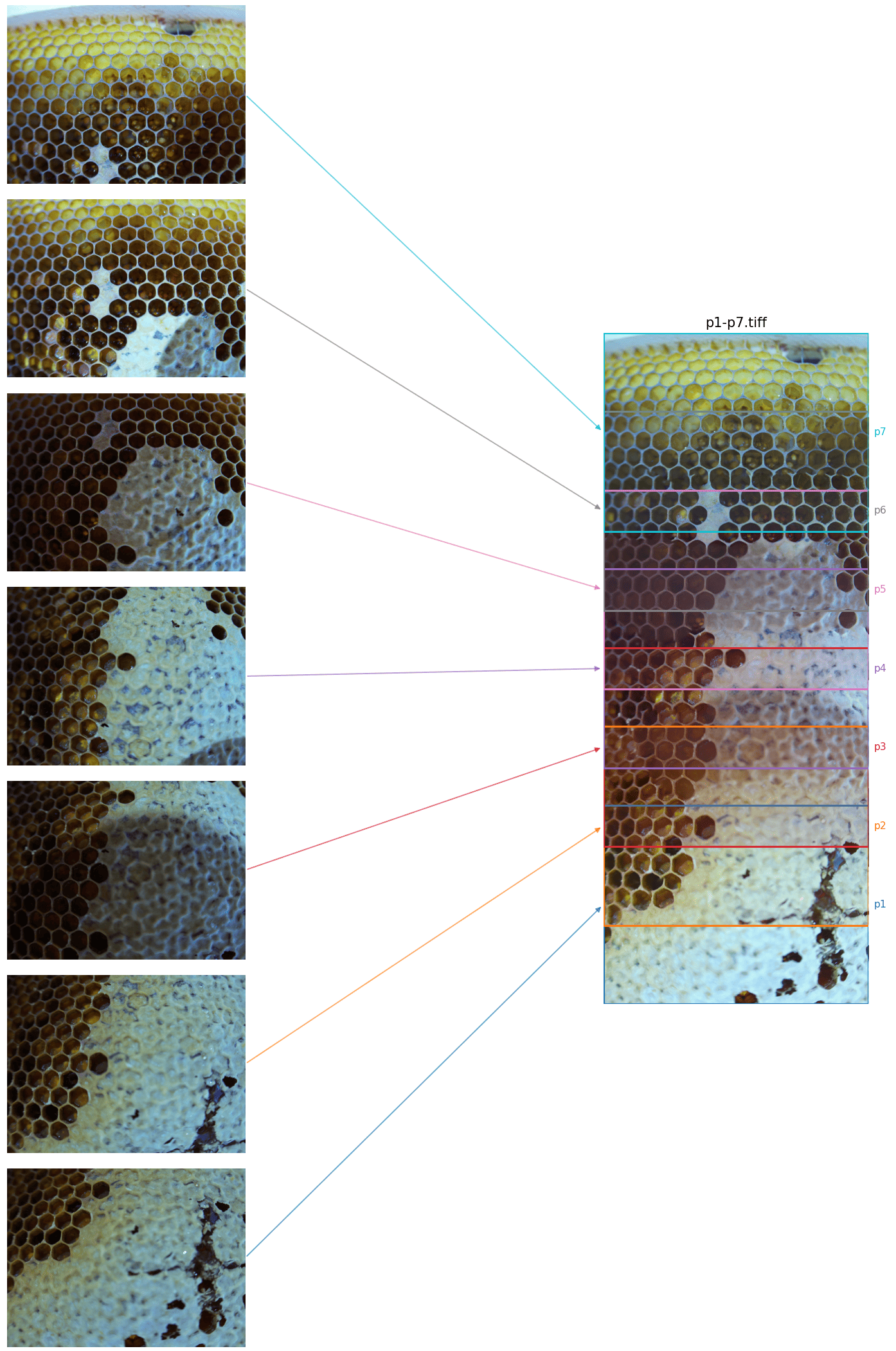}}
\caption{Vertical comb image constructed from seven image vertical pass}
\label{fig:figscanstitch}
\end{figure}

For the representative scan shown in Fig.~\ref{fig:figscanstitch}, the acquisition pattern consisted of seven vertically stacked scan rows and eight horizontal image positions per row. The vertical coverage was obtained with a nominal overlap of 60.4\% between successive rows, while the horizontal coverage used eight panels with a nominal overlap of 55.5\%. This overlap was selected to provide sufficient feature continuity for reliable stitching while still maintaining a practical scan duration. The average transition time between successive vertical scan positions was approximately 10.5~s, indicating that the platform can traverse between scan rows with consistent timing. This result demonstrates that the same robotic base used for signaling can also provide repeatable positioning for structured image acquisition.

\begin{figure}[htbp]
\centering
{\includegraphics[width=0.5\columnwidth]{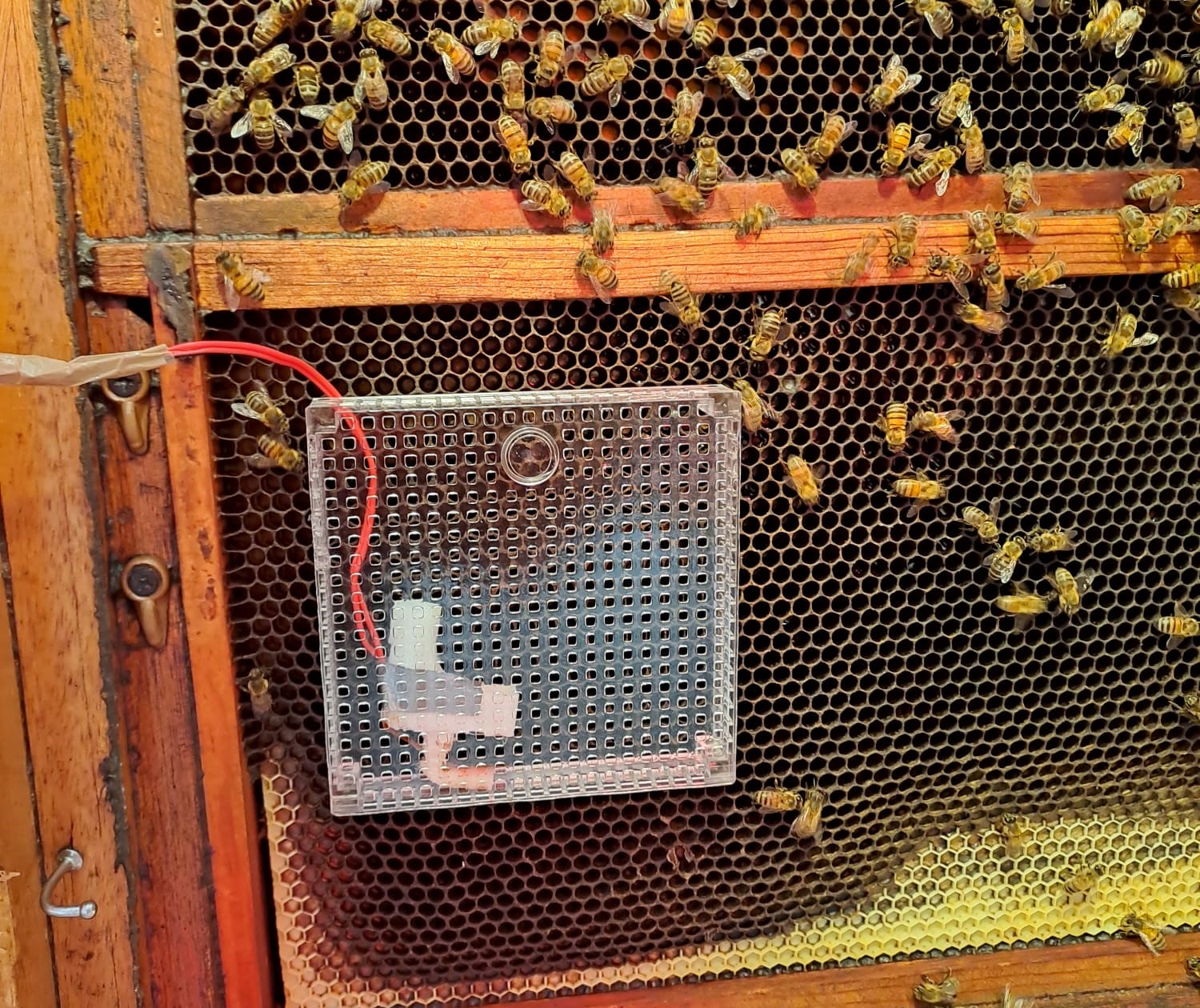}}
\caption{Bee dummy and flapper fixed to the comb for hive-scent acquisition}
\label{fig:figcage}
\end{figure}

\paragraph*{Oscillatory Signal Characterization} The electronic wing actuator was evaluated as a localized signaling payload capable of generating repeatable oscillatory motion at the comb interface. The dummy-bee payload and flapper assembly used for deployment and hive-scent acquisition are shown in Fig.~\ref{fig:figcage}. Rather than requiring full kinematic reconstruction, the output frequency of the flapper was estimated from video using a line-scan analysis of the region of strongest apparent motion. For each operating condition, a fixed line was placed through the wing trajectory, frame-wise intensity changes along that line were converted into a temporal signal, and the dominant frequency component was extracted from its spectrum.

\begin{figure}[htbp]
\centering
\subfigure[ \label{fig:13hzwing}]{
\includegraphics[width=\columnwidth]{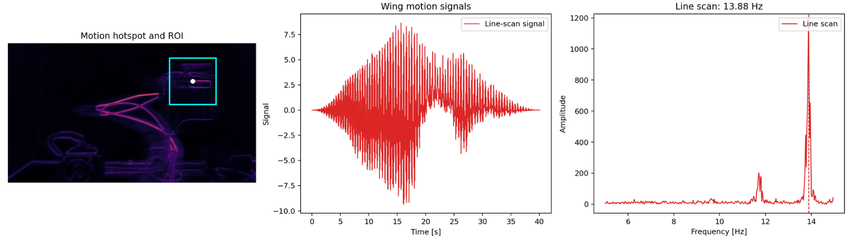}
}
\subfigure[ \label{fig:28hzwing} ]{\includegraphics[width=\columnwidth]{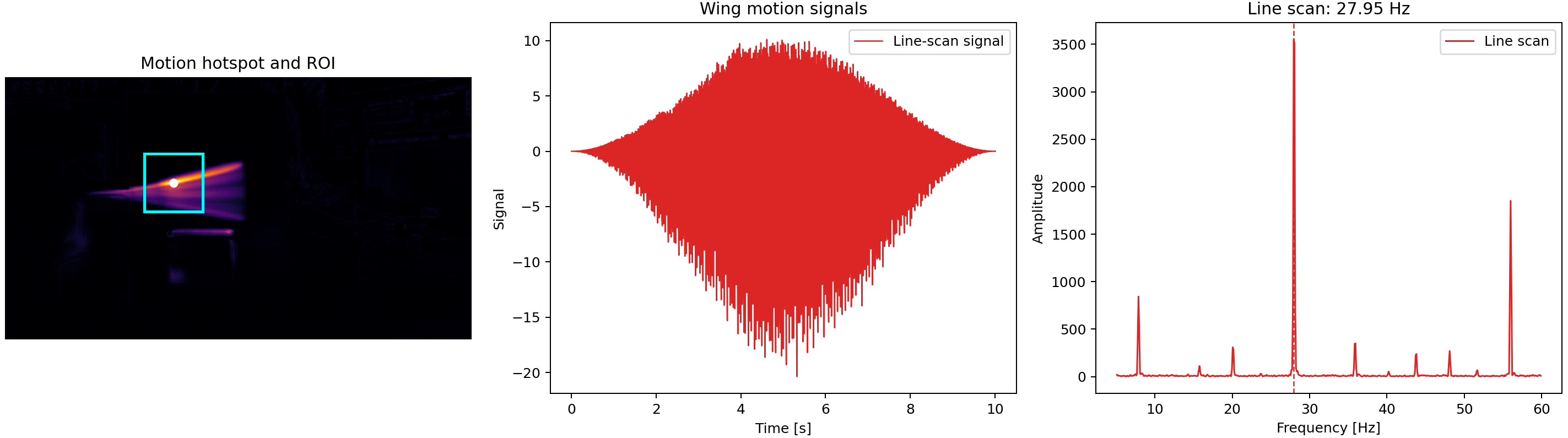}}
\caption{Wing-actuator frequency analysis (a) 13~Hz (b) 28~Hz.}
\end{figure}

In a low-frequency operating mode, the actuator exhibited a dominant oscillation at 13.88~Hz, as shown in Fig.~\ref{fig:13hzwing}. This value lies within the reported range of honeybee waggle- and shaking-signal frequencies and therefore demonstrates that the module can operate in a biologically relevant low-frequency regime. In a higher 28~Hz drive condition, the dominant oscillation increased to 27.95~Hz, as shown in Fig.~\ref{fig:28hzwing}, indicating that the same actuator architecture also provides substantial headroom for faster periodic motion beyond the shaking-like regime.

From a systems perspective, this module demonstrates that \textit{COMB} is not limited to carrying passive sensors or executing translational motion, but can also support compact local actuators whose outputs can be characterized quantitatively and synchronized with platform motion.

\paragraph*{In-Hive Operability and Maintenance} A final aspect of the evaluation concerned the practical operability of the platform in a live hive environment. During seasonal deployment, the Movable Access Window (MAW) remained functional despite propolis accumulation at the barrier interface. In practice, the MAW required one cleaning during the middle of the dance season and one replacement per season. This indicates that propolis fouling was present but manageable, and that the interface can support extended in-hive use with modest maintenance.

The hive-facing modules also proved physically compatible with repeated deployment. The final polycarbonate MAW and compliant dummy construction did not provoke obvious defensive responses during routine operation, and the platform remained usable over repeated sessions without major intervention beyond cleaning and part replacement at seasonal timescales. These observations do not constitute a behavioral validation of the signaling payloads, but they do establish that the system can be physically deployed and maintained inside a live colony over experimentally relevant periods.\par

Together, these results demonstrate that the complete system (stage, MAW, and end-effector) can deliver precise 2D dance trajectories, maintain hive compatibility over extended use, and actuate a localized signal (wing flapping) suitable for subsequent animal–robot interaction experiments.\par

\section{Conclusion and Future Work}
This paper presented \textit{COMB}, a compact, open-source, modular mechatronic platform for experimental access to honeybee colonies in standard observation-hive formats. Rather than introducing a single task-specific device, \textit{COMB} was designed as a shared robotic base that supports multiple sensing and actuation payloads within a common hive-compatible architecture. The platform combines a DNM-sized XY positioning stage, a Movable Access Window for repeated sealed access through the hive boundary, an embedded control stack, and interchangeable payload modules for observation and interaction tasks.

The system was evaluated in engineering terms through representative modules and deployment-oriented metrics. Video-based trajectory analysis showed that the platform can execute programmed waggle-like motion with millimeter-scale repeatability across repeated runs. In scanner mode, the same motion base supported tiled close-range image acquisition and stitched comb mosaics, demonstrating that the platform can also function as a repeatable in-hive imaging tool. The electronic wing module produced measurable oscillatory output in both a biologically relevant low-frequency regime and a higher-drive regime, confirming that the platform can support compact local actuation in addition to positioning and imaging. Finally, seasonal in-hive operation showed that the MAW and hive-facing modules remained usable with manageable maintenance under propolis exposure. Taken together, these results establish \textit{COMB} as a reusable mechatronic platform for controlled in-hive sensing and actuation.

Several directions remain for future work. On the hardware side, extending the platform beyond planar motion toward full three-dimensional comb-relative interaction would allow better accommodation of uneven comb geometry and enable more realistic motion primitives for biomimetic studies. The scanner mode can be strengthened through more systematic registration and repeated full-frame mosaicking for longer-term timelapse analysis. On the interaction side, future payloads could incorporate automated trophallaxis-style liquid delivery, localized sensing such as microphones, electric-field probes, or environmental sensors, and additional manipulators for probing or stimulation at selected comb locations. More broadly, the platform opens a path toward closed-loop in-hive experiments in which sensing, actuation, and behavioral tracking are integrated in a common system. By reducing the need for one-off hardware redesigns, \textit{COMB} aims to support more reproducible robotic experiments in honeybee research and related biohybrid systems.

\section*{Supplementary Materials}
All design artifacts required to reproduce our platform are openly archived on Zenodo at \url{https://zenodo.org/records/13693195}. The code is also made publicly available on GitHub at \url{ https://github.com/praked/COMB}.

\section*{Acknowledgment}
We thank K. Alomari at FU Berlin for their design suggestions, M. Stefanec at Uni Graz for help with scanner experiments, and James Nieh at UC Davis for permission to write about their bee-shaking signal experiment. This work was supported by the EU H2020 FET project HIVEOPOLIS (no. 824069).

\bibliographystyle{IEEEtran}
\bibliography{IEEEabrv,paper}

\vspace{12pt}
\color{red}

\end{document}